\documentclass[a4paper,conference]{IEEEtran}
\usepackage{cite}
\ifCLASSINFOpdf
\else
\fi
\usepackage{hyperref}

\usepackage{xcolor}
\usepackage{amsfonts, bm}       % blackboard math symbols
\usepackage{algorithm}
\usepackage{algorithmic} 
\usepackage{graphicx}
\usepackage{booktabs}       % professional-quality tables
\usepackage{tablefootnote}

\makeatletter
\usepackage[symbol]{footmisc}

\renewcommand{\thefootnote}{\fnsymbol{footnote}}

\newcommand{\printfnsymbol}[1]{%
  \textsuperscript{\@fnsymbol{#1}}%
}

\newcommand\blfootnote[1]{%
  \begingroup
  \renewcommand\thefootnote{}\footnote{#1}%
  \addtocounter{footnote}{-1}%
  \endgroup
}
\makeatother

\begin{document}
\title{Discovery of New Multi-Level Features for Domain Generalization via Knowledge Corruption}

\author{\IEEEauthorblockN{Ahmed Frikha}
\IEEEauthorblockA{Siemens Technology\\
University of Munich (LMU)\\
% Munich, Germany\\
ahmed.frikha@siemens.com}
\and
\IEEEauthorblockN{Denis Krompa{\ss}}
\IEEEauthorblockA{Siemens Technology\\
% Munich, Germany\\
denis.krompass@siemens.com}
\and
\IEEEauthorblockN{Volker Tresp}
\IEEEauthorblockA{Siemens Technology\\
University of Munich (LMU)\\
% Munich, Germany\\
volker.tresp@siemens.com}}


%Telephone: (800) 555--1212, Fax: (888) 555--1212}


\maketitle

\begin{abstract}
Machine learning models that can generalize to unseen domains are essential when applied in real-world scenarios involving strong domain shifts. We address the challenging domain generalization (DG) problem, where a model trained on a set of source domains is expected to generalize well in unseen domains without any exposure to their data. The main challenge of DG is that the features learned from the source domains are not necessarily present in the unseen target domains, leading to performance deterioration. We assume that learning a richer set of features is crucial to improve the transfer to a wider set of unknown domains. For this reason, we propose COLUMBUS, a method that enforces new feature discovery via a targeted corruption of the most relevant input and multi-level representations of the data. We conduct an extensive empirical evaluation to demonstrate the effectiveness of the proposed approach which achieves new state-of-the-art results by outperforming 18 DG algorithms on multiple DG benchmark datasets in the \textsc{DomainBed} framework.
\end{abstract}


% \fi
\IEEEpeerreviewmaketitle

\section{Introduction}

Deep learning models have achieved tremendous success when applied to independent and identically distributed (i.i.d.) data. However, in real-world applications, distribution shifts between training and test data are commonly encountered. For instance, data distributions might differ from one hospital to another \cite{dou2019domain}, and from one production plant to another. Similarly, the models in self-driving cars are exposed to different urban and rural environments in different countries with changing weather conditions \cite{yue2019domain} and object poses \cite{alcorn2019strike}.

\begin{figure}[h]
  \centering
  \includegraphics[width=0.47\textwidth]{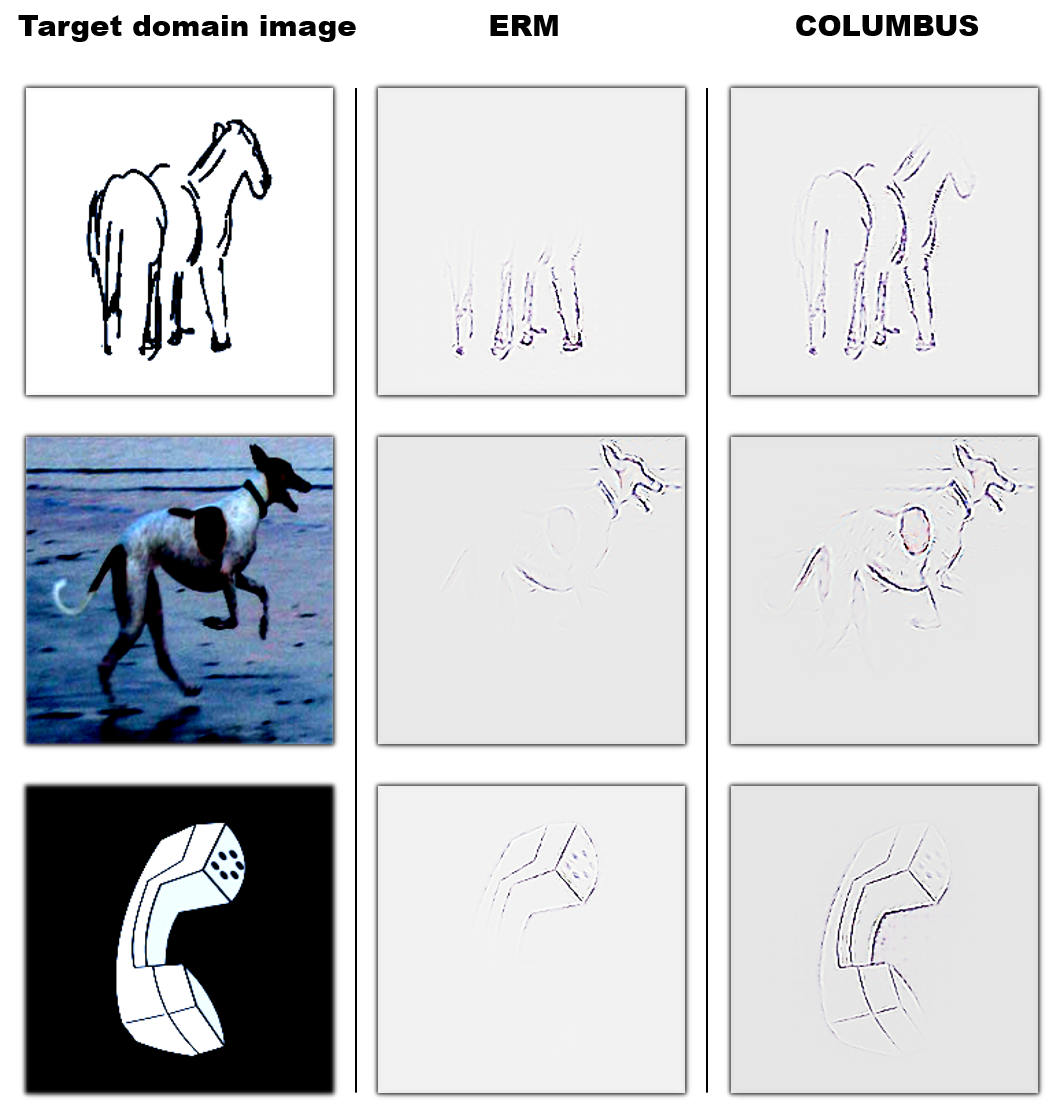}
  \caption{Relevance maps computed with GuidedGrad-CAM \cite{selvaraju2017grad} for ERM and COLUMBUS using images from the target domains, PACS \textit{Sketch}, VLCS \textit{VOC} and OfficeHome \textit{Clipart}. COLUMBUS recognizes more features than ERM, including horse back and muzzle, dog legs and tail, and phone shape.}
  \label{fig:ggcam}
\end{figure}
% Different scenarios of domain shifts were extensively studied in the literature. | Making ML models resiliennt to these domain drifts has been extensively studied in the literature.
Approaches to make machine learning models resilient to such data distribution changes were studied for different domain shift settings. For example, several domain adaptation methods were developed to address the case where, besides the data from the source domain(s), a set of labeled \cite{wang2018deep} or unlabeled \cite{wilson2020survey} data is available from a specific target domain. However, in real-world scenarios, collecting data from the target domain(s) is often slow, e.g., a new hospital or production site, expensive, or even infeasible, e.g., collecting images from every street of every country in the context of self-driving cars. Sometimes, the target domains cannot be known beforehand.

% However, a set of data examples from the target domain(s) is not always available in real-world scenarios. This may be due to expensive data collection, cold start situations where data can only be gradually collected, e.g., a new hospital or production site, or the infeasibility of data collection from every target domain, e.g., collecting images from every street of every country in the context of self-driving cars. In some situations, the target domains cannot be known at the time of the model training and deployment.
The Domain Generalization (DG) problem \cite{blanchard2011generalizing, muandet2013domain} was introduced to address such cases. Specifically, a model trained on multiple source domains is expected to directly perform well in unseen target domains without requiring any exposure to its data. This problem setting can be interpreted as multi-source 0-shot domain adaptation.

% CAN ADD: humans are good in DG 
Training a model to generalize across several related but unseen data distributions remains arguably one of the most challenging open problems in machine learning. In the last decade, a plethora of widely different methods were developed to address the DG problem. We refer to \cite{zhou2021domain} for an extensive overview of DG algorithms. Despite these efforts, \cite{gulrajani2020search} found that carefully tuning the baseline, which simply applies Empirical Risk Minimization (ERM) on the data of the source domains, achieves a high performance that is competitive with state-of-the-art methods. 

One major challenge of DG is that the model can only observe and learn features from the source domains, which may not be present in the unseen target domains, limiting generalization. We presume that learning a wider set of different features would increase the chance of learning features that are useful for a larger set of unseen domains. Hence, we introduce COLUMBUS, a training procedure for automated new feature discovery, which leads to a better feature recognition in unseen domains (Figure \ref{fig:ggcam}). During training on the source domains, COLUMBUS incentivizes the model to discover new features, even in data examples on which it already performs well. To achieve this, COLUMBUS prevents the model from using the features it deems most relevant for the source domains by corrupting them during training. To identify the most relevant features for a model, we leverage attribution methods \cite{tjoa2020survey} usually used for model explainability purposes.

% This is done via the identification and corruption of the features that are most relevant for the model's current prediction (Figure \ref{fig:method}). We apply this selective and model-state-specific corruption to the raw input data as well as its multi-level representations, and train the model on the corrupted data to promote a gradual new feature discovery. Hereby, the gradual learning of new features, independently from each other, promotes feature disentanglement. For the identification of the relevant features to be corrupted, we leverage attribution methods \cite{tjoa2020survey} usually used for model explainability purposes, such as GuidedGrad-CAM \cite{selvaraju2017grad}. 

We evaluate our approach on the recently proposed \textsc{DomainBed} framework \cite{gulrajani2020search} which includes several DG datasets and algorithm implementations to promote a fair and reproducible comparison of different approaches. Our method outperforms 18 DG algorithms evaluated on 3 datasets in the \textsc{DomainBed} framework, achieving new state-of-the-art results using 2 different model selection methods. Furthermore, our method achieves the highest performance when evaluated on unseen data from the source domains used for training (in-domain generalization), which confirms its effectiveness and ability to learn new features useful for unseen data.

\section{Related Work}
\label{related-work}

\subsection{Domain Generalization}

This section presents an overview of domain generalization (DG) approaches. Methods to which we compare in our experiments (Section \ref{exps}) are highlighted in bold. We refer to \cite{zhou2021domain} for an extensive overview of DG algorithms. The simplest approach to DG is to train one model via Empirical Risk Minimization (\textbf{ERM}) \cite{vapnik1999overview} on the training datasets of all source domains. \textbf{GroupDRO} \cite{sagawa2019distributionally} additionally increases the importance of source domains where the model yields a lower performance. In the following, we broadly categorize DG approaches into three categories.

\textbf{Domain alignment} methods aim to learn domain-invariant representations of the data by aligning features across the source domains. The reduction of the representation distribution mismatch across source domains can be achieved by minimizing the maximum mean discrepancy criteria \cite{gretton2012kernel} combined with an adversarial autoencoder (\textbf{MMD}) \cite{li2018domain}, minimizing the difference between the means \cite{tzeng2014deep} or covariance matrices (\textbf{CORAL}) \cite{sun2016deep} in the embedding space across different domains, or minimizing a contrastive loss \cite{motiian2017unified, yoon2019generalizable, mahajan2020domain}, e.g., \textbf{SelfReg} \cite{kim2021selfreg}. Domain alignment is also performed by aligning the loss gradients across source domains via inner product maximization (\textbf{Fish}) \cite{shi2021gradient}, or binary (\textbf{AND-mask}) \cite{parascandolo2020learning, shahtalebi2021sand} or continuous gradient masking (\textbf{SAND-mask}) \cite{shahtalebi2021sand}. 

Another line of works optimizes for features that confuse a domain discriminator model \cite{albuquerque2019generalizing, shao2019multi, rahman2020correlation, deng2020representation}, and includes \textbf{DANN} \cite{ganin2016domain} and its class-conditional extension \textbf{C-DANN} \cite{li2018deep}. Other works additionally involve the classifier in the representation alignment, either by optimizing for an embedding space such that the optimal linear classifier on top of it is the same across different domains (\textbf{IRM}) \cite{arjovsky2019invariant}, or by passing a domain-specific mean embedding to the classifier as a second argument (\textbf{MTL}) \cite{blanchard2017domain}. \textbf{VREx} \cite{krueger2021out} is an approximation of IRM via a variance penalty and \textbf{ARM} \cite{zhang2020adaptive} is an extension of MTL that employs a separate embedding CNN.

\textbf{Meta-learning} techniques were applied to DG by training a model in a bi-level optimization scheme on meta-train and meta-test sets sampled from the source domains. Hereby, \textbf{MLDG} \cite{li2018learning} optimizes for parameters that can be quickly adapted to different domains, MASF \cite{dou2019domain} adds inter-class and intra-class losses to regularize the embedding space, and MetaReg meta-learns a regularizer for the output layer \cite{balaji2018metareg}.

\textbf{Data augmentation} approaches were proposed to tackle DG and our method falls into this category. Some works use \textbf{Mixup} \cite{zhang2017mixup} to compute inter-domain examples to augment the training set \cite{xu2020adversarial, yan2020improve, wang2020heterogeneous}. \textbf{SagNets} \cite{nam2019reducing} reduce the domain gap by randomizing the style of images while keeping their content. Another line of works generate images by using adversarial attacks \cite{goodfellow2014explaining} to perturb input images based on a class classifier \cite{sinha2017certifying, volpi2018generalizing, qiao2020learning} or a domain classifier \cite{shankar2018generalizing}, by training CNNs to generate images within the source domains \cite{rahman2019multi, somavarapu2020frustratingly, borlino2021rethinking} or novel domains \cite{maria2019hallucinating, zhou2020deep, zhou2020learning}. Other works apply such perturbations on a feature level \cite{huang2020self, zhou2021mixstyle}. 

Our approach corrupts the raw input data as well as the multi-level representations that the model learns in order to enforce new feature discovery. Instead of using visually undetectable adversarial attacks or highly parametrized generative models, we employ attribution methods, e.g., Guided-Grad-CAM \cite{selvaraju2017grad}, to identify and corrupt the most relevant features. Our approach shares similarities with \textbf{RSC} \cite{huang2020self} which discards the most dominant features fed to the output layer to promote the activation of the remaining features. The key difference of our approach is that we corrupt features not only in the last high-level representation space, i.e., the input to the output layer, but also in the raw input space and other low-level representation spaces. We argue that by discarding the features only in the representation space (e.g., elephant trunk detector), as done in RSC, the same silenced feature detectors can be relearned as long as the model is exposed to the corresponding features in the input space (e.g., the pixels of the elephant trunk). We hypothesize that corrupting the features in the input space is crucial to enforce the discovery of new features. Our empirical results show that our method outperforms RSC by a significant margin (Section \ref{exps}) on unseen data from source and target domains, hence confirming our hypothesis.

\subsection{Relevance Attribution}
\label{rw-relevance}

In an attempt to explain and interpret the predictions of deep learning models, several attribution methods that assign relevance scores to input features have been developed \cite{simonyan2014very, selvaraju2017grad, schulz2020restricting}. In Saliency Maps \cite{simonyan2014very} the relevance scores are given by the gradient of the output neuron corresponding to the ground truth w.r.t. the input. Better attributions were achieved by averaging these gradients over local neighborhood patches in SmoothGrad \cite{smilkov2017smoothgrad} and over brightness level interpolations in IntegratedGradients \cite{sundararajan2017axiomatic}. Another category of approaches modifies the backpropagation procedure by considering only positive gradients \cite{springenberg2014striving} or to satisfy the relevance conservation property through the layers \cite{bach2015pixel, montavon2017explaining}. Class Activation Maps (CAM) \cite{zhou2016learning} leverages the activations in the last convolutional layer to produce a heatmap highlighting the relevance of each feature in the raw input. Gradient-weighted CAM (Grad-CAM) \cite{selvaraju2017grad} generalizes CAM to a variety of CNNs by using the gradient information flowing into the last convolutional layer. This method can be combined with GuidedBP \cite{springenberg2014striving} to yield GuidedGrad-CAM \cite{selvaraju2017grad}. IBA \cite{schulz2020restricting} approximates attribution scores by restricting the information flow via noise injection to intermediate feature maps during the forward pass.

While prior works used attribution methods to explain and interpret model predictions, we leverage them for training purposes. To the best of our knowledge, we are the first to incorporate attribution methods combined with data corruption into training to improve the model's generalization ability. For a broader overview of attribution methods, we refer to \cite{tjoa2020survey}. 

\section{Method}

The proposed method improves the knowledge transfer to unknown data distributions by training a model to learn a rich set of features on several representation levels of the data via an automated new feature discovery.

\begin{figure}[h]
  \centering
  \includegraphics[width=0.47\textwidth]{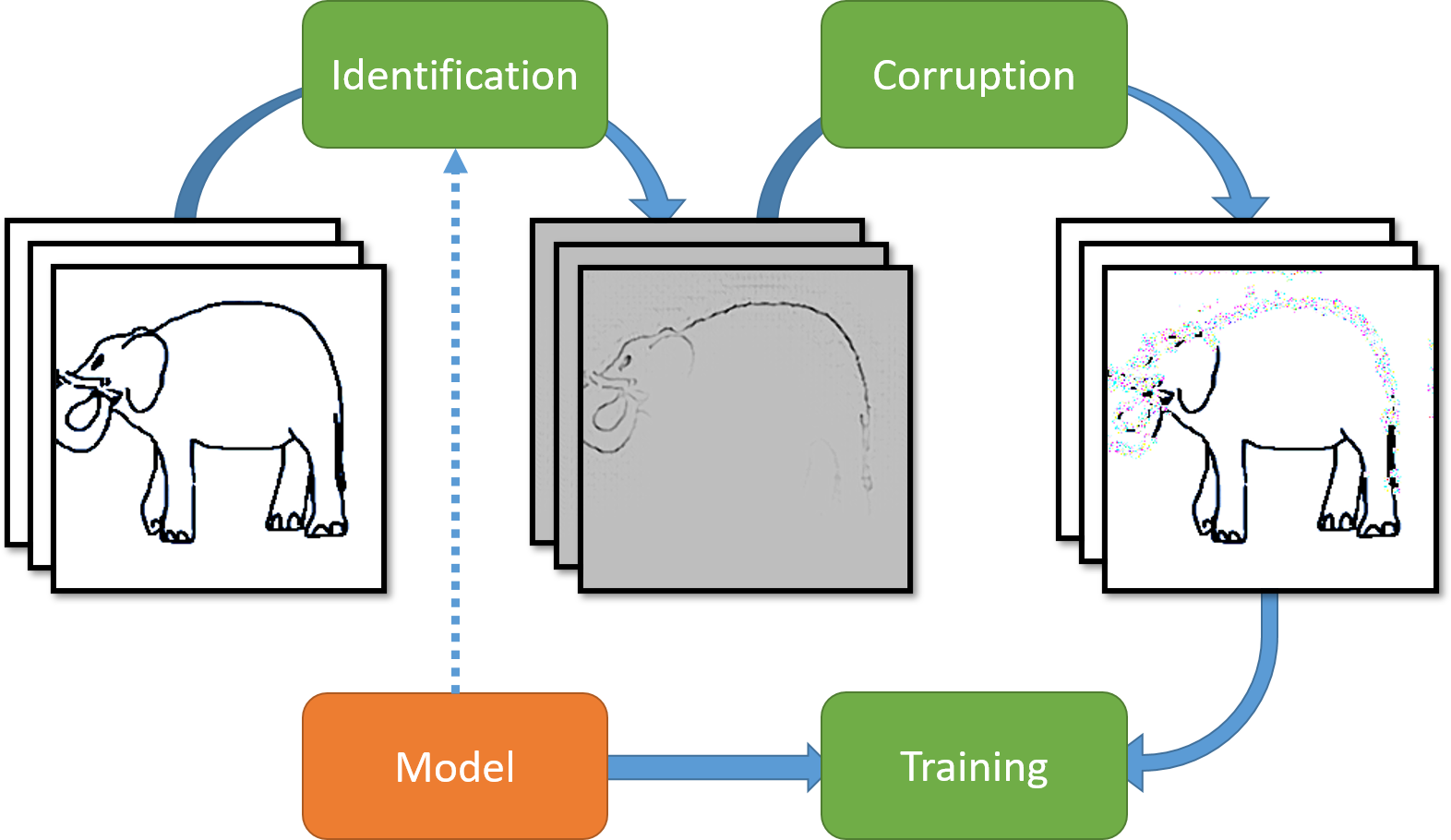}
  \caption{Overview of the proposed COLUMBUS method. In the identification stage, the most class-discriminative features according to the current model are identified via a relevance attribution method, which in this case is applied to the raw input representation. In the corruption stage, the identified features, e.g., elephant trunk and back, are perturbed by using a corruption method, in this case a replacement by a random pixel. Finally, the model is trained with the batch of corrupted data, promoting the discovery of new features, e.g., elephant feet and toes. The image used belongs to the PACS Sketch domain.}
  \label{fig:method}
\end{figure}

Let $F_{s}$ and $F_{t}$ denote the sets of features learnable for the addressed classification task, which are present in the data of the source domains and in the target domain, respectively, and $G$ their intersection. In the optimal case, the set of features $L$ learned by the model on the source domains encompasses $G$ fully. Since $F_{t}$ is unknown at training time, our method maximizes the size of $L$ by training the model to learn as many features as possible, resulting in a higher chance to capture features from $G$ via the higher intersection between $L$ and $G$. To achieve this, we propose COLUMBUS, a training procedure that enables automated new feature discovery. COLUMBUS prevents the model from using (a part of) the most relevant features for its current predictions during training. This is done in 3 major steps: identification of the most relevant features, their corruption, and training with the corrupted data representations. Figure \ref{fig:method} presents an overview of our approach. We apply this technique on several levels of representations of the data ranging from the raw input, e.g., pixels of the elephant trunk, to the high-level features fed to the output layers, e.g., elephant-trunk-detector, including the representations yielded by intermediate layers, hence fostering multi-level new feature discovery.
\subsection{Identification} In each training iteration, we sample a method from a set of attribution methods $A$, and use it to compute an attribution map $M$ that identifies the most relevant features. Any attribution method can be included in the set $A$. In this work, we use Saliency Maps \cite{simonyan2014very} and GuidedGrad-CAM \cite{selvaraju2017grad}, since they are simple, fast, and model-architecture-agnostic. Other methods require modifications to support skip connections and batch normalization layers \cite{bach2015pixel} or involve training additional parameters after each update \cite{schulz2020restricting}. Moreover, GuidedGrad-CAM was found to be competitive with the state-of-the-art attribution methods in the image degradation evaluation \cite{schulz2020restricting}.

While Saliency Maps and GuidedGrad-CAM were developed to assign relevance scores to features in the input space, we extend their usage to identify relevant features in representations extracted by intermediate layers. Let $\textbf{y}$ and $\textbf{\^y}$ denote the ground truth and the model prediction for a datapoint $\textbf{X}$. The attribution map $M_{l}$ yielded by Saliency Maps for a representation $R_{l}$ yielded by layer $l$ is given by
\begin{equation} \label{saliency}
    M_{l, Saliency} = \frac{\partial(\textbf{\^y}\odot\textbf{y})}{\partial R_{l}}.
\end{equation}
Note that the original Saliency Maps method \cite{simonyan2014very} corresponds to the case where $l=0$, i.e., $R_{0}$ is the raw input representation. 

The class-discriminative relevance map $M_{l}$ yielded by Grad-CAM \cite{selvaraju2017grad} for a layer $l$ is given by a sum over the channels of the representation $R_{l}$ weighted by importance factors $\alpha_{c}$ for each channel $c$, resulting in
\begin{equation} \label{ggcam}
    M_{l, Grad\-CAM} = ReLU(\sum_{c} \alpha_{c} R_{l}^{c}).
\end{equation}
Hereby, the importance factors are given by the gradient of the model prediction for the correct class w.r.t. the global-average-pooled representation $R_{l}$. Formally,  
\begin{equation} \label{ggcam_alpha}
    \alpha_{c} =\frac{1}{Z} \sum_{i}\sum_{j}\frac{\partial (\textbf{\^y}\odot\textbf{y})}{\partial R_{l}^{c,i,j}}.
\end{equation}
Grad-CAM is applied to the representation yielded by the last convolutional layer to obtain a relevance map $M$ which is upsampled to the input size \cite{selvaraju2017grad}. For intermediate representation $R_{l}$, we use the corresponding relevance map $M_{l}$ (Eq. \ref{ggcam}). We use GuidedGrad-CAM \cite{selvaraju2017grad} which yields more fine-grained maps than Grad-CAM by multiplying $M_{l, Grad\-CAM}$ with the relevance maps determined by Guided Backpropagation (Guided BP) \cite{springenberg2014striving}, for the same representation $R_{l}$. Guided BP modifies Saliency Maps by removing negative gradients when backpropagating through ReLU layers.
\subsection{Corruption} In each training iteration, we sample a method from the set of corruption methods $C$ and use it to corrupt the identified features based on the relevance attribution map $M$. Any technique that perturbs the information contained in the identified features can be used. We use different corruption methods depending on the sampled representation level $l$. 

To corrupt the raw input ($l=0$), the most relevant input features according to the attribution map $M_0$, e.g., the pixels corresponding to an elephant trunk, are perturbed using a corruption method. Hereby, we perturb the identified pixel values by setting them to a random value, to zero, i.e., black pixels, by applying the Fast Gradient Sign Method (FGSM) \cite{goodfellow2014explaining}, or by applying Gaussian blurring. For an intermediate representation level $l>0$, first the original input is fed through the model up to the corresponding layer $l$ to yield the representation $R_{l}$. The latter is then corrupted based on the relevance attribution map $M_l$ resulting from the identification stage and finally fed to the next layer. To corrupt intermediate embeddings, we drop the most relevant features, i.e., set their values to zero. This can be viewed as a targeted Dropout \cite{srivastava2014dropout}.

\subsection{Training} The COLUMBUS training procedure is described by Algorithm 1. In each training iteration, a data batch $B$, a representation level $l$, an attribution method, and a corruption method are sampled. Subsequently, the aforementioned identification and corruption steps are performed. The model is trained on the corrupted data (representations). Hereby, the $p\%$ most relevant features are corrupted. To enable the model to learn some features at the beginning of training, $p$ is set to 0, i.e., no corruption is applied. As training progresses, more features are corrupted in the data, forcing the model to discover and learn new features. Concretely, $p$ is linearly increased throughout the training to reach $p_{max}$, a hyperparameter. Note that the resulting gradual learning of new features, independently from each other, promotes also feature disentanglement, which was found to be beneficial for visual reasoning \cite{van2019disentangled}. We use different identification and corruption methods to increase the diversity of the corrupted datapoints used to train the model and prevent overfitting. We also found that sampling multiple methods leads to better empirical results. It should also be noted that COLUMBUS is adaptive to the model's learning, since the identification step is model-state-specific. In other words, if the model \textit{forgets} a set of features during training, these will not be identified (again), and hence will not be corrupted (again), which enables the model to relearn them.

The model parameters $\theta$ are updated by minimizing a loss function $f$ using a gradient-based optimization algorithm, e.g., Adam \cite{kingma2014adam}. In algorithm 1, SGD is used for simplicity of notation. The loss function $f$ used is a weighted sum comprising a classification loss $L_{cls}$ and a domain-alignment regularization loss term $L_{DA}$. Formally, 
\begin{equation} \label{loss_fct}
    f = L_{cls} + \lambda \sum_{i=1}^{N_s}\sum_{j=i+1}^{N_s}L_{DA}(i, j),
\end{equation}
\noindent where $\lambda$ is a weighting factor and $N_s$ the number of source domains. We use cross-entropy as classification loss $L_{cls}$. $L_{DA}$ regularizes the embedding space by minimizing the $l_2$-norm between the domain-specific embedding means and covariance matrices for each pair of source domains $i$ and $j$, as in DDC \cite{tzeng2014deep} and CORAL \cite{sun2016deep} respectively. The normalization of the regularization loss by the number of source domain pairs is omitted for simplicity of notation. We note that, unlike prior works \cite{tzeng2014deep, sun2016deep, gulrajani2020search} that apply this loss on the representations of the original images, we use corrupted images and representations. This leads to an alignment in the embedding space not only across the domains, but also between class-specific features, e.g., by minimizing the difference between the embedding distribution of cartoon images of elephant feet and art painting images of elephant trunks.

\begin{algorithm}
\caption{The COLUMBUS training procedure}
\label{algo}
\begin{algorithmic}[1]
\REQUIRE $D_{s}$: Training data of all source domains
\REQUIRE $f$: Loss function
\REQUIRE $\alpha$: Learning rate
\REQUIRE $L$: Set of representation levels including the raw input level
\REQUIRE $A$: Set of relevance attribution methods
\REQUIRE $C$: Set of corruption methods
\REQUIRE $p_{max}$: Max. \% of representation to be corrupted
  \STATE Initialize the model parameters $\theta$ randomly or from a pre-trained model
  \STATE Initialize the \% of representation to be corrupted $p=0$
    \WHILE{not done}
      \STATE Sample a data batch $B = \{\textbf{X}, \textbf{y}\}$ from $D_{s}$\\
      \STATE Sample level of representation $l$ from $L$ \\
      \STATE Feed $B$ through the model parametrized by $\theta$ \\
      \STATE Get the relevance attribution map $M_{l}$ by applying a relevance attribution method randomly sampled from $A$ on level $l$ using the current model parameters $\theta$\\
      \IF{$l=0$}
        \STATE Corrupt the values in $B$ corresponding to the $p$\% highest values in $M_{0}$ by applying a corruption method randomly sampled from $C$, yielding the corrupted input $B_{c}$\\
        \STATE Feed $B_{c}$ through the model to obtain the predictions $\textbf{\^y}$ \\
        % \STATE Get the model predictions $\textbf{\^y}$ for $B_{c}$  \\
      \ELSE
        \STATE Feed $B$ through the model until level $l$ to obtain the representation $R_{l}$\\
        \STATE Corrupt the values in $R_{l}$ corresponding to the $p$\% highest values in $M_{l}$, yielding the corrupted representation $R_{l, c}$\\
        \STATE Feed $R_{l, c}$ through the model starting from level $l+1$ to obtain the predictions $\textbf{\^y}$\\
      \ENDIF\\
      \STATE Update $\theta$: $\theta \gets \theta - \alpha \nabla_{\theta} f(\textbf{y}, \textbf{\^y})$\\
      \STATE Increase $p$ linearly towards $p_{max}$
    \ENDWHILE \\
\STATE \textbf{return} Learned model parameters $\theta$
\end{algorithmic}
\end{algorithm}

In our experiments, we corrupt $q\%$ of the sampled data batch in each iteration, and increase $q$ linearly during the training until $q_{max}$ is reached, as done for the representation percentage to be corrupted $p$. This is omitted in Algorithm 1 for simplicity of notation. During training, we alternate between sampling an intermediate representation and the raw input for corruption. The intermediate representations correspond to the outputs of each ResNet block in the used ResNet-50 model \cite{he2016deep}. At test time, the model trained with COLUMBUS is applied to the data from the target domains without any corruption.

\section{Experiments}
\label{exps}

\subsection{Experimental Setup}
\label{exp-setup}
% The conducted experiments\footnote{Code will be made public upon paper acceptance.} aim to address the following key questions: $(a)$ Can COLUMBUS achieve a higher DG performance than the simple but strong baseline ERM \cite{gulrajani2020search}, which simply trains the model on the source domains, and how does it compare to the state-of-the-art DG methods ? $(b)$ Does learning a richer set of features with COLUMBUS lead to a performance increase on unseen data from the source domains ? $(c)$ How much does the embeddings space regularization loss $L_{DA}$ impact the performance of COLUMBUS ?

We evaluate our approach empirically\footnote[1]{Code under \url{https://github.com/AhmedFrikha/columbus-domainbed}.} on the recently proposed \textsc{DomainBed} framework \cite{gulrajani2020search} which includes several DG datasets, implementations of DG algorithms, and model selection methods. \textsc{DomainBed} promotes a fair and reproducible comparison of the different approaches by including a common automated hyperparameter search, i.e., a random search with the given seeds conducts the same experiments for all methods. For a fair comparison with the 18 DG algorithms, our experiments follow the same experimental setting adopted in \textsc{DomainBed} \cite{gulrajani2020search}: We use a ResNet-50 model \cite{he2016deep} pre-trained on ImageNet \cite{russakovsky2015imagenet} with frozen batch normalization \cite{ioffe2015batch} statistics as suggested in \cite{seo2020learning}, the same optimization algorithm, data augmentation techniques and number of training iterations used in \textsc{DomainBed}. The COLUMBUS-specific hyperparameters $p_{max}$, $q_{max}$ and $\lambda$ are included in the hyperparameter search of \textsc{DomainBed}, and the intervals used can be found in the Appendix. We noticed that the published code \cite{gulrajani2020search} with the provided seeds does not enable the reproduction of the published results, since the resulting points in the hyperparameter search space are different from the ones used for the published results. Therefore, for a fairer comparison, we additionally rerun the experiments of the best performing DG method in \textsc{DomainBed}, i.e., CORAL, with the published code and seeds that we used for COLUMBUS.

We conduct experiments on 3 challenging multi-domain datasets commonly used as DG benchmarks: VLCS \cite{fang2013unbiased}, OfficeHome \cite{venkateswara2017deep} and PACS \cite{li2017deeper}. VLCS contains images belonging to 5 classes from 4 photographic domains: VOC2007 (V), LabelMe (L), Caltech101 (C), and SUN09 (S). OfficeHome consists of images of 65 classes from the domains Art (A), Clipart (C), Product (P), and Real (R). PACS comprises images belonging to 7 classes from the domains Art-painting (A), Cartoon (C), Photo (P), and Sketch (S). \textsc{DomainBed} splits each source domain data into $80\%$ for training and $20\%$ for validation. Each experiment is run with the provided 3 seeds.

\subsection{Results}
% In this section, we present the results of the experiments conducted to answer our questions (Section \ref{exp-setup}). First, we present the domain generalization results achieved by COLUMBUS and the baselines on \textsc{DomainBed}, using two different DG selection methods. Subsequently, we demonstrate the effectiveness of our approach on in-domain generalization. Finally, we assess the impact of removing the domain alignment regularization loss, by evaluating a \textit{non-regularized} version of our approach (NR-COLUMBUS).

% \subsubsection{Domain Generalization}
Tables \ref{r_vlcs}, \ref{r_pacs} and \ref{r_office} show the results averaged over the 3 seeds pre-determined by \textsc{DomainBed}, on VLCS, PACS and OfficeHome respectively. Hereby, the unseen target domain is defined by the column name, i.e., the 3 other domains are used as source domains for training. The test accuracy is computed on the test set of the target domain. We provide results including standard deviations in the appendix. The average results over the domains of each dataset can be seen in Table \ref{r_avg}. We select the model with the highest source-domain validation performance for the evaluation on the target domain.\blfootnote{\printfnsymbol{2}Results yielded by using published code \cite{gulrajani2020search} with the provided seeds.}

\begin{table}[h]
\caption{Domain Generalization results on VLCS.}
\label{r_vlcs}
\centering
\small
% \begin{center}
% \adjustbox{max width=0.47\textwidth}{%
\begin{tabular}{lccccc}
% \toprule
\textbf{Algorithm}   & \textbf{C}           & \textbf{L}           & \textbf{S}           & \textbf{V}           & \textbf{Avg}         \\
\midrule
ERM                  & 97.7        & 64.3        & 73.4        & 74.6        & 77.5                 \\
IRM                  & 98.6        & 64.9        & 73.4        & 77.3        & 78.5                 \\
GroupDRO             & 97.3        & 63.4        & 69.5        & 76.7        & 76.7                 \\
Mixup                & 98.3        & 64.8        & 72.1        & 74.3        & 77.4                 \\
MLDG                 & 97.4        & 65.2        & 71.0        & 75.3        & 77.2                 \\
CORAL                & 98.3        & 66.1        & 73.4        & 77.5        & 78.8                 \\
MMD                  & 97.7        & 64.0        & 72.8        & 75.3        & 77.5                 \\
DANN                 & 99.0        & 65.1        & 73.1        & 77.2        & 78.6                 \\
CDANN                & 97.1        & 65.1        & 70.7        & 77.1        & 77.5                 \\
MTL                  & 97.8        & 64.3        & 71.5        & 75.3        & 77.2                 \\
SagNet               & 97.9        & 64.5        & 71.4        & 77.5        & 77.8                 \\
ARM                  & 98.7        & 63.6        & 71.3        & 76.7        & 77.6                 \\
VREx                 & 98.4        & 64.4        & 74.1        & 76.2        & 78.3                 \\
RSC                  & 97.9        & 62.5        & 72.3        & 75.6        & 77.1                 \\
\midrule
% \small{NR-COLUMBUS}                 & 99.0        & 63.0        & 74.4        & 78.6        & 78.7                 \\
CORAL\printfnsymbol{2}              & 97.3       & 65.2       & 71.5       & 75.6       & 77.4 \\
COLUMBUS           & 98.9       & 65.0       & 75.0      & 77.9       & 79.2                 \\
% \bottomrule
\end{tabular}
% }
% \end{center}
\end{table}

\begin{table}[h]
\caption{Domain Generalization results on PACS.}
\label{r_pacs}
\centering
\small
% \begin{center}
% \adjustbox{max width=0.47\textwidth}{%
\begin{tabular}{lccccc}
% \toprule
\textbf{Algorithm}   & \textbf{A}           & \textbf{C}           & \textbf{P}           & \textbf{S}           & \textbf{Avg}         \\
\midrule
ERM                  & 84.7        & 80.8        & 97.2        & 79.3        & 85.5                 \\
IRM                  & 84.8        & 76.4        & 96.7        & 76.1        & 83.5                 \\
GroupDRO             & 83.5        & 79.1        & 96.7        & 78.3       & 84.4                 \\
Mixup                & 86.1        & 78.9        & 97.6        & 75.8        & 84.6                 \\
MLDG                 & 85.5        & 80.1        & 97.4        & 76.6        & 84.9                 \\
CORAL                & 88.3        & 80.0        & 97.5        & 78.8        & 86.2                 \\
MMD                  & 86.1        & 79.4        & 96.6        & 76.5        & 84.6                 \\
DANN                 & 86.4        & 77.4        & 97.3        & 73.5       & 83.6                 \\
CDANN                & 84.6        & 75.5        & 96.8        & 73.5        & 82.6                 \\
MTL                  & 87.5        & 77.1        & 96.4        & 77.3        & 84.6                 \\
SagNet               & 87.4        & 80.7        & 97.1        & 80.0        & 86.3                 \\
ARM                  & 86.8        & 76.8        & 97.4        & 79.3        & 85.1                 \\
VREx                 & 86.0       & 79.1        & 96.9        & 77.7        & 84.9                 \\
RSC                  & 85.4        & 79.7        & 97.6        & 78.2        & 85.2                 \\
\midrule
% \small{NR-COLUMBUS}                 & 88.3        & 79.3        & 97.3        & 81.5        & 86.6                 \\
CORAL\printfnsymbol{2}                   & 87.4       & 79.4       & 97.5       & 73.9       & 84.5 \\
COLUMBUS           & 88.7       & 78.7       & 97.2       & 81.5       & 86.5                 \\
% \bottomrule
\end{tabular}
% }
% \end{center}
\end{table}

\begin{table}[h]
% \begin{center}
\caption{Domain generalization results on OfficeHome.}
\label{r_office}
\centering
\small
% \adjustbox{max width=0.47\textwidth}{%
\begin{tabular}{lccccc}
% \toprule
\textbf{Algorithm}   & \textbf{A}           & \textbf{C}           & \textbf{P}           & \textbf{R}           & \textbf{Avg}         \\
\midrule
ERM                  & 61.3        & 52.4        & 75.8        & 76.6        & 66.5                 \\
IRM                  & 58.9       & 52.2        & 72.1       & 74.0        & 64.3                 \\
GroupDRO             & 60.4        & 52.7        & 75.0        & 76.0        & 66.0                 \\
Mixup                & 62.4        & 54.8        & 76.9        & 78.3        & 68.1                 \\
MLDG                 & 61.5        & 53.2        & 75.0        & 77.5        & 66.8                 \\
CORAL                & 65.3        & 54.4        & 76.5        & 78.4        & 68.7                 \\
MMD                  & 60.4        & 53.3        & 74.3        & 77.4        & 66.3                 \\
DANN                 & 59.9        & 53.0        & 73.6        & 76.9        & 65.9                 \\
CDANN                & 61.5        & 50.4        & 74.4        & 76.6        & 65.8                 \\
MTL                  & 61.5        & 52.4        & 74.9        & 76.8        & 66.4                 \\
SagNet               & 63.4        & 54.8        & 75.8        & 78.3        & 68.1                 \\
ARM                  & 58.9        & 51.0        & 74.1        & 75.2        & 64.8                 \\
VREx                 & 60.7        & 53.0        & 75.3        & 76.6        & 66.4                 \\
RSC                  & 60.7        & 51.4        & 74.8        & 75.1        & 65.5                 \\
\midrule
% \small{NR-COLUMBUS}                 & 61.6        & 58.4        & 75.4        & 77.5        & 68.2                 \\
CORAL\printfnsymbol{2}                  & 64.8       & 54.6       & 76.8       & 78.4       & 68.6 \\
COLUMBUS           & 62.8       & 57.9       & 75.5       & 77.9       & 68.5                 \\
% \bottomrule
\end{tabular}
% }
% \end{center}
\end{table}

\begin{table}[h]
\caption{Average domain generalization results.}
\label{r_avg}
\centering
% \adjustbox{max width=0.47\textwidth}{%
\begin{tabular}{lcccc}
% \toprule
\textbf{Algorithm}        & \textbf{VLCS}             & \textbf{PACS}             & \textbf{\small{OfficeHome}}       & \textbf{Avg}              \\
\midrule
ERM                       & 77.5 $\pm$ 0.4            & 85.5 $\pm$ 0.2            & 66.5 $\pm$ 0.3            & 76.5                      \\
IRM                       & 78.5 $\pm$ 0.5            & 83.5 $\pm$ 0.8            & 64.3 $\pm$ 2.2            & 75.5                      \\
GroupDRO                  & 76.7 $\pm$ 0.6            & 84.4 $\pm$ 0.8            & 66.0 $\pm$ 0.7            & 75.7                      \\
Mixup                     & 77.4 $\pm$ 0.6            & 84.6 $\pm$ 0.6            & 68.1 $\pm$ 0.3            & 76.7                      \\
MLDG                      & 77.2 $\pm$ 0.4            & 84.9 $\pm$ 1.0             & 66.8 $\pm$ 0.6            & 76.3                      \\
CORAL                     & 78.8 $\pm$ 0.6            & 86.2 $\pm$ 0.3            & 68.7 $\pm$ 0.3            & 77.9                      \\
MMD                       & 77.5 $\pm$ 0.9            & 84.6 $\pm$ 0.5            & 66.3 $\pm$ 0.1            & 76.2                      \\
DANN                      & 78.6 $\pm$ 0.4            & 83.6 $\pm$ 0.4            & 65.9 $\pm$ 0.6            & 76.0                      \\
CDANN                     & 77.5 $\pm$ 0.1            & 82.6 $\pm$ 0.9            & 65.8 $\pm$ 1.3            & 75.3                      \\
MTL                       & 77.2 $\pm$ 0.4            & 84.6 $\pm$ 0.5            & 66.4 $\pm$ 0.5            & 76.1                      \\
SagNet                    & 77.8 $\pm$ 0.5            & 86.3 $\pm$ 0.2            & 68.1 $\pm$ 0.1            & 77.4                      \\
ARM                       & 77.6 $\pm$ 0.3            & 85.1 $\pm$ 0.4            & 64.8 $\pm$ 0.3            & 75.8                      \\
VREx                      & 78.3 $\pm$ 0.2            & 84.9 $\pm$ 0.6            & 66.4 $\pm$ 0.6            & 76.5                      \\
RSC                       & 77.1 $\pm$ 0.5            & 85.2 $\pm$ 0.9            & 65.5 $\pm$ 0.9            & 75.9                      \\
SelfReg                   & 77.8 $\pm$ 0.9            & 85.6 $\pm$ 0.4            & 67.9 $\pm$ 0.7            & 77.1                      \\
% IIB                       & 78.6 $\pm$ 0.3            & 87.4 $\pm$ 0.7            & 66.9 $\pm$ 1.1            & 77.6                      \\
Fish                      & 77.8 $\pm$ 0.3            & 85.5 $\pm$ 0.3            & 68.6 $\pm$ 0.4            & 77.3                      \\
AND-mask                  & 78.1 $\pm$ 0.9            & 84.4 $\pm$ 0.9            & 65.6 $\pm$ 0.4            & 76.0                      \\
SAND-mask                 & 77.4 $\pm$ 0.2            & 84.6 $\pm$ 0.9            & 65.8 $\pm$ 0.4            & 75.9                      \\
\midrule
% \small{NR-COLUMBUS}                      & 78.7 $\pm$ 0.3            & 86.6 $\pm$ 0.1            & 68.2 $\pm$ 0.2            & 77.8                      \\
CORAL\printfnsymbol{2}                       & 77.4 $\pm$ 0.3            & 84.5 $\pm$ 0.5            & 68.6 $\pm$ 0.2            & 76.9 \\
COLUMBUS                & 79.2 $\pm$ 0.2            & 86.5 $\pm$ 0.4            & 68.5 $\pm$ 0.4            & 78.1                      \\
% \bottomrule
\end{tabular}
% }
% \end{center}

\end{table}

COLUMBUS achieves the highest results on all datasets on average, advancing the state-of-the-art by $1.6\%$ and $1.2\%$ compared to ERM and CORAL respectively. We note an impressive $5.5\%$ improvement on OfficeHome's most challenging domain \textit{Clipart} (\textbf{C}) compared to ERM and $3.3\%$ compared to CORAL, on this $65$-class classification task. Likewise, on the \textit{Art} (\textbf{A}) domain of PACS, substantial $4\%$ and $1.3\%$ increases are observed compared to ERM and CORAL respectively. A significant performance increase is achieved on PACS's challenging \textit{Sketch} (\textbf{S}) domain as well. On all target domains, COLUMBUS consistently outperforms all the baselines or yields a competitive performance. The fact that COLUMBUS outperforms RSC \cite{huang2020self} confirms our hypothesis, that corrupting the learned features in the raw input is crucial to prevent relearning the same high-level features, and hence enforce new feature discovery.

\begin{table}[h]
\caption{Domain Generalization results using the test-domain validation set (oracle) as a selection method.}
\label{r_oracle}
\centering
% \adjustbox{max width=0.47\textwidth}{%
\begin{tabular}{lcccc}
% \toprule
\textbf{Algorithm}        & \textbf{VLCS}             & \textbf{PACS}             & \textbf{\small{OfficeHome}}       & \textbf{Avg}              \\
\midrule
ERM                       & 77.6 $\pm$ 0.3            & 86.7 $\pm$ 0.3            & 66.4 $\pm$ 0.5            & 76.9                      \\
IRM                       & 76.9 $\pm$ 0.6            & 84.5 $\pm$ 1.1            & 63.0 $\pm$ 2.7            & 74.8                      \\
GroupDRO                  & 77.4 $\pm$ 0.5            & 87.1 $\pm$ 0.1            & 66.2 $\pm$ 0.6            & 76.9                      \\
Mixup                     & 78.1 $\pm$ 0.3            & 86.8 $\pm$ 0.3            & 68.0 $\pm$ 0.2            & 77.6                      \\
MLDG                      & 77.5 $\pm$ 0.1            & 86.8 $\pm$ 0.4            & 66.6 $\pm$ 0.3            & 77.0                      \\
CORAL                     & 77.7 $\pm$ 0.2            & 87.1 $\pm$ 0.5            & 68.4 $\pm$ 0.2            & 77.7                      \\
MMD                       & 77.9 $\pm$ 0.1            & 87.2 $\pm$ 0.1            & 66.2 $\pm$ 0.3            & 77.1                      \\
DANN                      & 79.7 $\pm$ 0.5            & 85.2 $\pm$ 0.2            & 65.3 $\pm$ 0.8            & 76.8                      \\
CDANN                     & 79.9 $\pm$ 0.2            & 85.8 $\pm$ 0.8            & 65.3 $\pm$ 0.5            & 77.0                      \\
MTL                       & 77.7 $\pm$ 0.5            & 86.7 $\pm$ 0.2            & 66.5 $\pm$ 0.4            & 77.0                      \\
SagNet                    & 77.6 $\pm$ 0.1            & 86.4 $\pm$ 0.4            & 67.5 $\pm$ 0.2            & 77.2                      \\
ARM                       & 77.8 $\pm$ 0.3            & 85.8 $\pm$ 0.2            & 64.8 $\pm$ 0.4            & 76.1                      \\
VREx                      & 78.1 $\pm$ 0.2            & 87.2 $\pm$ 0.6            & 65.7 $\pm$ 0.3            & 77.0                      \\
RSC                       & 77.8 $\pm$ 0.6            & 86.2 $\pm$ 0.5            & 66.5 $\pm$ 0.6            & 76.8                      \\
AND-mask                  & 76.4 $\pm$ 0.4            & 86.4 $\pm$ 0.4            & 66.1 $\pm$ 0.2            & 76.3                      \\
SAND-mask                 & 76.2 $\pm$ 0.5            & 85.9 $\pm$ 0.4            & 65.9 $\pm$ 0.5            & 76.0                      \\
\midrule
% \small{NR-COLUMBUS}                      & 79.2 $\pm$ 0.2            & 87.9 $\pm$ 0.3            & 68.4 $\pm$ 0.2            & 78.5                      \\
CORAL\printfnsymbol{2}                       & 77.4 $\pm$ 0.6            & 85.6 $\pm$ 0.8            & 68.4 $\pm$ 0.4            & 77.1                      \\
COLUMBUS                & 77.7 $\pm$ 0.4            & 88.2 $\pm$ 0.2            & 69.6 $\pm$ 0.4            & 78.5                      \\
% \bottomrule
\end{tabular}
\end{table}

We also evaluate our approach using the oracle selection method \cite{gulrajani2020search}, where the model is evaluated on a held-out validation set from the target domain. In order to limit access to the target domain, this evaluation is performed only once at the end of each training, disallowing early stopping. The average results are presented in Table \ref{r_oracle}. We find that the performance advantage of COLUMBUS is increased when better proxies for model selection, e.g., a held-out set from the target domain, are available, further confirming the effectiveness of our approach. Our results on \textsc{DomainBed} using both model selection methods show that the additional features learned thanks to the corruption of the most relevant features are useful for generalization to unseen domains. This is backed by Figure \ref{fig:ggcam}, where COLUMBUS recognizes more features in examples from the unseen target domain than ERM.

% Finally, we evaluate COLUMBUS and the DG baselines on the held-out validation sets of the source domains to investigate their in-domain generalization. The results (provided in the Appendix) show that COLUMBUS consistently achieves the highest validation performance on the training domains compared to the DG baselines, on every dataset. Hence, the richer set of learned features also improves generalization to unseen in-distribution data examples, i.e., from the source domains, suggesting that COLUMBUS might be suitable for applications beyond domain generalization to include scenarios without domain shift.

Finally, we investigate whether the richer set of features learned by COLUMBUS leads to a better in-domain generalization, i.e., whether a performance boost is also yielded on unseen source domain data. We evaluate COLUMBUS and the DG baselines on the held-out validation sets of the source domains and report the maximal mean validation accuracy across domains in the Appendix. COLUMBUS consistently achieves the highest validation performance on the training domains compared to the DG baselines. This shows that the richer set of learned features improves generalization to unseen in-distribution datapoints, suggesting that COLUMBUS might also be suitable for applications without domain shift.

% \subsubsection{Ablation Study}
% Finally, we conduct a small ablation study to assess the impact of the domain alignment regularization loss term $L_{DA}$. For that, we evaluate a \textit{non-regularized} version of COLUMBUS, i.e., $\lambda=0$, to which refer as NR-COLUMBUS, in the same experimental setting. The results can be seen in Tables 1-6. 

% We find that, on average, the regularization yields a performance increase for in-domain and out-of-domain generalization, using both selection methods. Only on VLCS, it leads to a marginal ($0.1\%$) performance decrease. In comparison to the DG baselines, NR-COLUMBUS still outperforms all the other methods in in-domain generalization and domain generalization using the oracle selection method.

\section{Conclusion}
In this work, we proposed COLUMBUS, a novel and strong domain generalization (DG) approach that enforces new feature discovery to improve the transfer to a wider set of unseen domains. During training, COLUMBUS corrupts the input and multi-level representations of the data most relevant for the model. For the identification of such features, relevance attribution methods that are usually used for model explainability purposes are leveraged. Our extensive empirical evaluation on \textsc{DomainBed} demonstrates the effectiveness of the proposed method, which outperforms 18 DG algorithms and achieves new state-of-the-art results on multiple DG benchmarks. Our results show that the richer set of learned features improves the generalization to unseen data from both seen and unseen domains, suggesting the suitability of our approach for applications beyond domain generalization to include scenarios without domain shift.

% \cleardoublepage

% \cleardoublepage
\appendix

\textbf{Experimental Setting Details}
In this section we provide further details about the experiments conducted. The experiments were conducted on computing instances that include a Tesla T4 NVIDIA GPU, 8 custom Intel Cascade Lake CPUs and 32 Gb of memory. The operating system used is Ubuntu 20.04 LTS. The libraries PyTorch \cite{paszke2019pytorch} and TorchVision were used with the versions 1.7.1 and 0.8.2, respectively. 

In our experiments, the percentage of representation corrupted $p$ and the percentage of the batch corrupted $q$ are increased linearly towards $p_{max}$ and $q_{max}$, respectively, during the first half of the training. In the second half of the training, the maximum values are used. 
% While the gradual learning of new features happens in the first half of the training, we view the second half as a consolidation 

For a fair comparison, we used the automated hyperparameter search from \textsc{DomainBed} \cite{gulrajani2020search} for each domain and dataset. Hereby, each hyperparameter search involves 20 random search experiments, i.e., the hyperparameters are randomly sampled from the specified intervals. To distribute the hyperparameter search experiments over multiple devices (each experiment runs on a single GPU), we used the Ray Tune package \cite{moritz2018ray, liaw2018tune}. Our experiments follow the experimental setting: We use a ResNet-50 model \cite{he2016deep} pretrained on ImageNet \cite{russakovsky2015imagenet} with frozen batch normalization \cite{ioffe2015batch} statistics as suggested in \cite{seo2020learning}, as well as the same optimization algorithm, ADAM \cite{kingma2014adam}, data augmentation techniques, and number of training iterations. An overview of the hyperparameter-specific intervals we used for COLUMBUS can be seen in Table \ref{hps}. The algorithm-specific hyperparameter intervals used for the other DG algorithms can be found in \cite{gulrajani2020search}. Depending on whether the corruption is applied on the input or an intermediate representation, different value intervals were used for the percentage of the representation corrupted $p$ and the percentage of the batch corrupted $q$. For the hyperparameters related to intermediate representations, i.e., $p_{max, intermediate}$ and $q_{max, intermediate}$, the interval upper bounds were chosen based on the results of RSC \cite{huang2020self}, which discards the most dominant features fed to the output layer, i.e., the last representation level. We used the same intervals used in \textsc{DomainBed} for the other algorithms for all hyperparameters. \\

\begin{table*}[h]
\caption{Hyperparameter intervals used for the hyperparameter search conducted with \textsc{DomainBed}}
\label{hps}
\begin{center}
% \adjustbox{max width=0.47\textwidth}{%
\begin{tabular}{ll}
% \toprule
\textbf{Hyperparameter}   & \textbf{Random Distribution} \\
\midrule
Weighting coefficient $\lambda$                  & $10^{Uniform(-1,1)}$\\
Max. corruption \% for input representation $p_{max,input}$                 & $Uniform(0.2,0.5)$\\
Max. corruption \% for intermediate representation $p_{max,intermediate}$                 & $Uniform(0.01,0.333)$\\
Max. batch corruption \% for input representation $q_{max,input}$                 & $Uniform(0.2,1.0)$\\
Max. batch corruption \% for intermediate representation $q_{max,intermediate}$                 & $Uniform(0.1,0.5)$\\
% \bottomrule
\end{tabular}
% }
\end{center}
\end{table*}

\textbf{Source Domain Generalization}

In this section, we investigate whether the richer set of features learned by COLUMBUS leads to a better in-domain generalization, i.e., whether a performance boost is also yielded on unseen data from the source domains used for training. We evaluate COLUMBUS and the DG baselines on the held-out validation sets of the source domains and report the maximal average validation accuracy across domains in Table \ref{r_val}\footnote{For the baselines, we computed the results using the logs made public in \url{https://drive.google.com/file/d/16VFQWTble6-nB5AdXBtQpQFwjEC7CChM/view?usp=sharing}.}.

\begin{table*}[h]
\caption{Source domain validation performance.}
\label{r_val}
\centering
% \begin{center}
% \adjustbox{max width=0.47\textwidth}{%
\begin{tabular}{lcccc}
% \toprule
\textbf{Algorithm}        & \textbf{VLCS}             & \textbf{PACS}             & \textbf{\small{OfficeHome}}       & \textbf{Avg}              \\
\midrule
ERM                       & 86.4 $\pm$ 0.0            & 97.0 $\pm$ 0.1            & 82.1 $\pm$ 0.2            & 88.5                      \\
IRM                       & 85.8 $\pm$ 0.2            & 96.5 $\pm$ 0.4            & 79.9 $\pm$ 2.0            & 87.4                      \\
GroupDRO                  & 86.4 $\pm$ 0.0            & 96.9 $\pm$ 0.1            & 81.6 $\pm$ 0.2            & 88.3                      \\
Mixup                     & 86.6 $\pm$ 0.1            & 97.4 $\pm$ 0.1            & 83.2 $\pm$ 0.3            & 89.0                      \\
MLDG                      & 86.4 $\pm$ 0.1            & 97.1 $\pm$ 0.1            & 82.4 $\pm$ 0.3            & 88.6                      \\
CORAL                     & 86.5 $\pm$ 0.0            & 97.1 $\pm$ 0.1            & 83.7 $\pm$ 0.2            & 89.1                      \\
MMD                       & 86.4 $\pm$ 0.1            & 96.9 $\pm$ 0.0            & 82.0 $\pm$ 0.1            & 88.4                      \\
DANN                      & 86.3 $\pm$ 0.0            & 96.4 $\pm$ 0.3            & 80.4 $\pm$ 0.9            & 87.7                      \\
CDANN                     & 86.4 $\pm$ 0.1            & 96.4 $\pm$ 0.3            & 80.5 $\pm$ 0.9            & 87.8                      \\
MTL                       & 86.3 $\pm$ 0.0            & 97.0 $\pm$ 0.0            & 81.7 $\pm$ 0.2            & 88.3                      \\
SagNet                    & 86.4 $\pm$ 0.0            & 97.0 $\pm$ 0.2            & 82.9 $\pm$ 0.4            & 88.8                      \\
ARM                       & 86.3 $\pm$ 0.0            & 96.5 $\pm$ 0.1            & 80.2 $\pm$ 0.2            & 87.7                      \\
VREx                      & 86.2 $\pm$ 0.1            & 96.9 $\pm$ 0.1            & 81.8 $\pm$ 0.4            & 88.3                      \\
RSC                       & 86.4 $\pm$ 0.3            & 96.8 $\pm$ 0.2            & 81.5 $\pm$ 0.3            & 88.2                      \\
\midrule
CORAL\printfnsymbol{2}                     & 86.6 $\pm$ 0.1            & 96.8 $\pm$ 0.2            & 83.6 $\pm$ 0.0            & 89.0                      \\
COLUMBUS                & 86.6 $\pm$ 0.1            & 97.3 $\pm$ 0.0            & 83.4 $\pm$ 0.1            & 89.1                      \\
% \bottomrule
\end{tabular}
% }
% \end{center}
\end{table*}

COLUMBUS consistently achieves the highest validation performance on the training domains compared to the DG baselines, on every dataset. This shows that the richer set of learned features improves generalization to unseen in-distribution data examples as well, suggesting that COLUMBUS might be suitable for applications beyond domain generalization to include scenarios without domain shift.\\

\textbf{Results including standard deviations}
We present the domain generalization results of COLUMBUS and the baselines, including the standard deviations computed over the 3 runs with the seeds provided by \textsc{DomainBed} in Tables \ref{r_vlcs_std}, \ref{r_pacs_std} and \ref{r_office_std}. Hereby, for model selection, the \textit{training-domain validation-set} from \textsc{DomainBed} is used. 

\begin{table*}[t]
\caption{Domain Generalization results on VLCS, including standard deviation.}
\label{r_vlcs_std}
\begin{center}
% \adjustbox{max width=0.47\textwidth}{%
\begin{tabular}{lccccc}
% \toprule
\textbf{Algorithm}   & \textbf{C}           & \textbf{L}           & \textbf{S}           & \textbf{V}           & \textbf{Avg}         \\
\midrule
ERM                  & 97.7 $\pm$ 0.4       & 64.3 $\pm$ 0.9       & 73.4 $\pm$ 0.5       & 74.6 $\pm$ 1.3       & 77.5                 \\
IRM                  & 98.6 $\pm$ 0.1       & 64.9 $\pm$ 0.9       & 73.4 $\pm$ 0.6       & 77.3 $\pm$ 0.9       & 78.5                 \\
GroupDRO             & 97.3 $\pm$ 0.3       & 63.4 $\pm$ 0.9       & 69.5 $\pm$ 0.8       & 76.7 $\pm$ 0.7       & 76.7                 \\
Mixup                & 98.3 $\pm$ 0.6       & 64.8 $\pm$ 1.0       & 72.1 $\pm$ 0.5       & 74.3 $\pm$ 0.8       & 77.4                 \\
MLDG                 & 97.4 $\pm$ 0.2       & 65.2 $\pm$ 0.7       & 71.0 $\pm$ 1.4       & 75.3 $\pm$ 1.0       & 77.2                 \\
CORAL                & 98.3 $\pm$ 0.1       & 66.1 $\pm$ 1.2       & 73.4 $\pm$ 0.3       & 77.5 $\pm$ 1.2       & 78.8                 \\
MMD                  & 97.7 $\pm$ 0.1       & 64.0 $\pm$ 1.1       & 72.8 $\pm$ 0.2       & 75.3 $\pm$ 3.3       & 77.5                 \\
DANN                 & 99.0 $\pm$ 0.3       & 65.1 $\pm$ 1.4       & 73.1 $\pm$ 0.3       & 77.2 $\pm$ 0.6       & 78.6                 \\
CDANN                & 97.1 $\pm$ 0.3       & 65.1 $\pm$ 1.2       & 70.7 $\pm$ 0.8       & 77.1 $\pm$ 1.5       & 77.5                 \\
MTL                  & 97.8 $\pm$ 0.4       & 64.3 $\pm$ 0.3       & 71.5 $\pm$ 0.7       & 75.3 $\pm$ 1.7       & 77.2                 \\
SagNet               & 97.9 $\pm$ 0.4       & 64.5 $\pm$ 0.5       & 71.4 $\pm$ 1.3       & 77.5 $\pm$ 0.5       & 77.8                 \\
ARM                  & 98.7 $\pm$ 0.2       & 63.6 $\pm$ 0.7       & 71.3 $\pm$ 1.2       & 76.7 $\pm$ 0.6       & 77.6                 \\
VREx                 & 98.4 $\pm$ 0.3       & 64.4 $\pm$ 1.4       & 74.1 $\pm$ 0.4       & 76.2 $\pm$ 1.3       & 78.3                 \\
RSC                  & 97.9 $\pm$ 0.1       & 62.5 $\pm$ 0.7       & 72.3 $\pm$ 1.2       & 75.6 $\pm$ 0.8       & 77.1                 \\
\midrule
CORAL\printfnsymbol{2}                 & 97.3 $\pm$ 0.3       & 65.2 $\pm$ 0.5       & 71.5 $\pm$ 0.6       & 75.6 $\pm$ 0.9       & 77.4                 \\
COLUMBUS           & 98.9 $\pm$ 0.2       & 65.0 $\pm$ 1.3       & 75.0 $\pm$ 0.2       & 77.9 $\pm$ 0.9       & 79.2                 \\
% \bottomrule
\end{tabular}
% }
\end{center}
\end{table*}

\begin{table*}[t]
\caption{Domain Generalization results on PACS, including standard deviation.}
\label{r_pacs_std}
\begin{center}
% \adjustbox{max width=0.47\textwidth}{%
\begin{tabular}{lccccc}
% \toprule
\textbf{Algorithm}   & \textbf{A}           & \textbf{C}           & \textbf{P}           & \textbf{S}           & \textbf{Avg}         \\
\midrule
ERM                  & 84.7 $\pm$ 0.4       & 80.8 $\pm$ 0.6       & 97.2 $\pm$ 0.3       & 79.3 $\pm$ 1.0       & 85.5                 \\
IRM                  & 84.8 $\pm$ 1.3       & 76.4 $\pm$ 1.1       & 96.7 $\pm$ 0.6       & 76.1 $\pm$ 1.0       & 83.5                 \\
GroupDRO             & 83.5 $\pm$ 0.9       & 79.1 $\pm$ 0.6       & 96.7 $\pm$ 0.3       & 78.3 $\pm$ 2.0       & 84.4                 \\
Mixup                & 86.1 $\pm$ 0.5       & 78.9 $\pm$ 0.8       & 97.6 $\pm$ 0.1       & 75.8 $\pm$ 1.8       & 84.6                 \\
MLDG                 & 85.5 $\pm$ 1.4       & 80.1 $\pm$ 1.7       & 97.4 $\pm$ 0.3       & 76.6 $\pm$ 1.1       & 84.9                 \\
CORAL                & 88.3 $\pm$ 0.2       & 80.0 $\pm$ 0.5       & 97.5 $\pm$ 0.3       & 78.8 $\pm$ 1.3       & 86.2                 \\
MMD                  & 86.1 $\pm$ 1.4       & 79.4 $\pm$ 0.9       & 96.6 $\pm$ 0.2       & 76.5 $\pm$ 0.5       & 84.6                 \\
DANN                 & 86.4 $\pm$ 0.8       & 77.4 $\pm$ 0.8       & 97.3 $\pm$ 0.4       & 73.5 $\pm$ 2.3       & 83.6                 \\
CDANN                & 84.6 $\pm$ 1.8       & 75.5 $\pm$ 0.9       & 96.8 $\pm$ 0.3       & 73.5 $\pm$ 0.6       & 82.6                 \\
MTL                  & 87.5 $\pm$ 0.8       & 77.1 $\pm$ 0.5       & 96.4 $\pm$ 0.8       & 77.3 $\pm$ 1.8       & 84.6                 \\
SagNet               & 87.4 $\pm$ 1.0       & 80.7 $\pm$ 0.6       & 97.1 $\pm$ 0.1       & 80.0 $\pm$ 0.4       & 86.3                 \\
ARM                  & 86.8 $\pm$ 0.6       & 76.8 $\pm$ 0.5       & 97.4 $\pm$ 0.3       & 79.3 $\pm$ 1.2       & 85.1                 \\
VREx                 & 86.0 $\pm$ 1.6       & 79.1 $\pm$ 0.6       & 96.9 $\pm$ 0.5       & 77.7 $\pm$ 1.7       & 84.9                 \\
RSC                  & 85.4 $\pm$ 0.8       & 79.7 $\pm$ 1.8       & 97.6 $\pm$ 0.3       & 78.2 $\pm$ 1.2       & 85.2                 \\
\midrule
CORAL\printfnsymbol{2}                 & 87.4 $\pm$ 0.3       & 79.4 $\pm$ 0.3       & 97.5 $\pm$ 0.1       & 73.9 $\pm$ 1.8       & 84.5                 \\
COLUMBUS           & 88.7 $\pm$ 0.8       & 78.7 $\pm$ 1.0       & 97.2 $\pm$ 0.1       & 81.5 $\pm$ 1.5       & 86.5                 \\
% \bottomrule
\end{tabular}
% }
\end{center}
\end{table*}

\begin{table*}[ht]
\caption{Domain generalization results on OfficeHome, including standard deviation.}
\label{r_office_std}
\begin{center}
% \adjustbox{max width=0.47\textwidth}{%
\begin{tabular}{lccccc}
\toprule
\textbf{Algorithm}   & \textbf{A}           & \textbf{C}           & \textbf{P}           & \textbf{R}           & \textbf{Avg}         \\
\midrule
ERM                  & 61.3 $\pm$ 0.7       & 52.4 $\pm$ 0.3       & 75.8 $\pm$ 0.1       & 76.6 $\pm$ 0.3       & 66.5                 \\
IRM                  & 58.9 $\pm$ 2.3       & 52.2 $\pm$ 1.6       & 72.1 $\pm$ 2.9       & 74.0 $\pm$ 2.5       & 64.3                 \\
GroupDRO             & 60.4 $\pm$ 0.7       & 52.7 $\pm$ 1.0       & 75.0 $\pm$ 0.7       & 76.0 $\pm$ 0.7       & 66.0                 \\
Mixup                & 62.4 $\pm$ 0.8       & 54.8 $\pm$ 0.6       & 76.9 $\pm$ 0.3       & 78.3 $\pm$ 0.2       & 68.1                 \\
MLDG                 & 61.5 $\pm$ 0.9       & 53.2 $\pm$ 0.6       & 75.0 $\pm$ 1.2       & 77.5 $\pm$ 0.4       & 66.8                 \\
CORAL                & 65.3 $\pm$ 0.4       & 54.4 $\pm$ 0.5       & 76.5 $\pm$ 0.1       & 78.4 $\pm$ 0.5       & 68.7                 \\
MMD                  & 60.4 $\pm$ 0.2       & 53.3 $\pm$ 0.3       & 74.3 $\pm$ 0.1       & 77.4 $\pm$ 0.6       & 66.3                 \\
DANN                 & 59.9 $\pm$ 1.3       & 53.0 $\pm$ 0.3       & 73.6 $\pm$ 0.7       & 76.9 $\pm$ 0.5       & 65.9                 \\
CDANN                & 61.5 $\pm$ 1.4       & 50.4 $\pm$ 2.4       & 74.4 $\pm$ 0.9       & 76.6 $\pm$ 0.8       & 65.8                 \\
MTL                  & 61.5 $\pm$ 0.7       & 52.4 $\pm$ 0.6       & 74.9 $\pm$ 0.4       & 76.8 $\pm$ 0.4       & 66.4                 \\
SagNet               & 63.4 $\pm$ 0.2       & 54.8 $\pm$ 0.4       & 75.8 $\pm$ 0.4       & 78.3 $\pm$ 0.3       & 68.1                 \\
ARM                  & 58.9 $\pm$ 0.8       & 51.0 $\pm$ 0.5       & 74.1 $\pm$ 0.1       & 75.2 $\pm$ 0.3       & 64.8                 \\
VREx                 & 60.7 $\pm$ 0.9       & 53.0 $\pm$ 0.9       & 75.3 $\pm$ 0.1       & 76.6 $\pm$ 0.5       & 66.4                 \\
RSC                  & 60.7 $\pm$ 1.4       & 51.4 $\pm$ 0.3       & 74.8 $\pm$ 1.1       & 75.1 $\pm$ 1.3       & 65.5                 \\
\midrule
CORAL\printfnsymbol{2}                 & 64.8 $\pm$ 0.2       & 54.6 $\pm$ 0.7       & 76.8 $\pm$ 0.6       & 78.4 $\pm$ 0.3       & 68.6                 \\
COLUMBUS           & 62.8 $\pm$ 0.3       & 57.9 $\pm$ 0.8       & 75.5 $\pm$ 0.1       & 77.9 $\pm$ 0.5       & 68.5                 \\
\bottomrule
\end{tabular}
% }
\end{center}

\end{table*}

%\section*{Acknowledgment}




%\bibliographystyle{IEEEtran}


% \end{thebibliography}
\cleardoublepage

\cleardoublepage
\bibliographystyle{IEEEtran}
\bibliography{references}

\begin{thebibliography}{10}
\providecommand{\url}[1]{#1}
\csname url@samestyle\endcsname
\providecommand{\newblock}{\relax}
\providecommand{\bibinfo}[2]{#2}
\providecommand{\BIBentrySTDinterwordspacing}{\spaceskip=0pt\relax}
\providecommand{\BIBentryALTinterwordstretchfactor}{4}
\providecommand{\BIBentryALTinterwordspacing}{\spaceskip=\fontdimen2\font plus
\BIBentryALTinterwordstretchfactor\fontdimen3\font minus
  \fontdimen4\font\relax}
\providecommand{\BIBforeignlanguage}[2]{{%
\expandafter\ifx\csname l@#1\endcsname\relax
\typeout{** WARNING: IEEEtran.bst: No hyphenation pattern has been}%
\typeout{** loaded for the language `#1'. Using the pattern for}%
\typeout{** the default language instead.}%
\else
\language=\csname l@#1\endcsname
\fi
#2}}
\providecommand{\BIBdecl}{\relax}
\BIBdecl

\bibitem{dou2019domain}
Q.~Dou, D.~Coelho~de Castro, K.~Kamnitsas, and B.~Glocker, ``Domain
  generalization via model-agnostic learning of semantic features,''
  \emph{NeurIPS}, 2019.

\bibitem{yue2019domain}
X.~Yue, Y.~Zhang, S.~Zhao, A.~Sangiovanni-Vincentelli, K.~Keutzer, and B.~Gong,
  ``Domain randomization and pyramid consistency: Simulation-to-real
  generalization without accessing target domain data,'' in \emph{IEEE/CVF
  ICCV}, 2019.

\bibitem{alcorn2019strike}
M.~A. Alcorn, Q.~Li, Z.~Gong, C.~Wang, L.~Mai, W.-S. Ku, and A.~Nguyen,
  ``Strike (with) a pose: Neural networks are easily fooled by strange poses of
  familiar objects,'' in \emph{IEEE/CVF CVPR}, 2019.

\bibitem{selvaraju2017grad}
R.~R. Selvaraju, M.~Cogswell, A.~Das, R.~Vedantam, D.~Parikh, and D.~Batra,
  ``Grad-cam: Visual explanations from deep networks via gradient-based
  localization,'' in \emph{IEEE ICCV}, 2017.

\bibitem{wang2018deep}
M.~Wang and W.~Deng, ``Deep visual domain adaptation: A survey,''
  \emph{Neurocomputing}, 2018.

\bibitem{wilson2020survey}
G.~Wilson and D.~J. Cook, ``A survey of unsupervised deep domain adaptation,''
  \emph{ACM Transactions on Intelligent Systems and Technology (TIST)}, 2020.

\bibitem{blanchard2011generalizing}
G.~Blanchard, G.~Lee, and C.~Scott, ``Generalizing from several related
  classification tasks to a new unlabeled sample,'' \emph{NeurIPS}, 2011.

\bibitem{muandet2013domain}
K.~Muandet, D.~Balduzzi, and B.~Sch{\"o}lkopf, ``Domain generalization via
  invariant feature representation,'' in \emph{ICML}, 2013.

\bibitem{zhou2021domain}
K.~Zhou, Z.~Liu, Y.~Qiao, T.~Xiang, and C.~C. Loy, ``Domain generalization: A
  survey,'' \emph{arXiv:2103.02503}, 2021.

\bibitem{gulrajani2020search}
I.~Gulrajani and D.~Lopez-Paz, ``In search of lost domain generalization,''
  \emph{arXiv:2007.01434}, 2020.

\bibitem{tjoa2020survey}
E.~Tjoa and C.~Guan, ``A survey on explainable artificial intelligence (xai):
  Toward medical xai,'' \emph{IEEE Transactions on Neural Networks and Learning
  Systems}, 2020.

\bibitem{vapnik1999overview}
V.~N. Vapnik, ``An overview of statistical learning theory,'' \emph{IEEE
  transactions on neural networks}, 1999.

\bibitem{sagawa2019distributionally}
S.~Sagawa, P.~W. Koh, T.~B. Hashimoto, and P.~Liang, ``Distributionally robust
  neural networks for group shifts: On the importance of regularization for
  worst-case generalization,'' \emph{arXiv:1911.08731}, 2019.

\bibitem{gretton2012kernel}
A.~Gretton, K.~M. Borgwardt, M.~J. Rasch, B.~Sch{\"o}lkopf, and A.~Smola, ``A
  kernel two-sample test,'' \emph{JMLR}, 2012.

\bibitem{li2018domain}
H.~Li, S.~J. Pan, S.~Wang, and A.~C. Kot, ``Domain generalization with
  adversarial feature learning,'' in \emph{IEEE CVPR}, 2018.

\bibitem{tzeng2014deep}
E.~Tzeng, J.~Hoffman, N.~Zhang, K.~Saenko, and T.~Darrell, ``Deep domain
  confusion: Maximizing for domain invariance,'' \emph{arXiv:1412.3474}, 2014.

\bibitem{sun2016deep}
B.~Sun and K.~Saenko, ``Deep coral: Correlation alignment for deep domain
  adaptation,'' in \emph{ECCV}, 2016.

\bibitem{motiian2017unified}
S.~Motiian, M.~Piccirilli, D.~A. Adjeroh, and G.~Doretto, ``Unified deep
  supervised domain adaptation and generalization,'' in \emph{IEEE ICCV}, 2017.

\bibitem{yoon2019generalizable}
C.~Yoon, G.~Hamarneh, and R.~Garbi, ``Generalizable feature learning in the
  presence of data bias and domain class imbalance with application to skin
  lesion classification,'' in \emph{International Conference on Medical Image
  Computing and Computer-Assisted Intervention}, 2019.

\bibitem{mahajan2020domain}
D.~Mahajan, S.~Tople, and A.~Sharma, ``Domain generalization using causal
  matching,'' \emph{arXiv:2006.07500}, 2020.

\bibitem{kim2021selfreg}
D.~Kim, S.~Park, J.~Kim, and J.~Lee, ``Selfreg: Self-supervised contrastive
  regularization for domain generalization,'' \emph{arXiv:2104.09841}, 2021.

\bibitem{shi2021gradient}
Y.~Shi, J.~Seely, P.~H. Torr, N.~Siddharth, A.~Hannun, N.~Usunier, and
  G.~Synnaeve, ``Gradient matching for domain generalization,''
  \emph{arXiv:2104.09937}, 2021.

\bibitem{parascandolo2020learning}
G.~Parascandolo, A.~Neitz, A.~Orvieto, L.~Gresele, and B.~Sch{\"o}lkopf,
  ``Learning explanations that are hard to vary,'' \emph{arXiv:2009.00329},
  2020.

\bibitem{shahtalebi2021sand}
S.~Shahtalebi, J.-C. Gagnon-Audet, T.~Laleh, M.~Faramarzi, K.~Ahuja, and
  I.~Rish, ``Sand-mask: An enhanced gradient masking strategy for the discovery
  of invariances in domain generalization,'' \emph{arXiv:2106.02266}, 2021.

\bibitem{albuquerque2019generalizing}
I.~Albuquerque, J.~Monteiro, M.~Darvishi, T.~H. Falk, and I.~Mitliagkas,
  ``Generalizing to unseen domains via distribution matching,''
  \emph{arXiv:1911.00804}, 2019.

\bibitem{shao2019multi}
R.~Shao, X.~Lan, J.~Li, and P.~C. Yuen, ``Multi-adversarial discriminative deep
  domain generalization for face presentation attack detection,'' in
  \emph{IEEE/CVF CVPR}, 2019.

\bibitem{rahman2020correlation}
M.~M. Rahman, C.~Fookes, M.~Baktashmotlagh, and S.~Sridharan,
  ``Correlation-aware adversarial domain adaptation and generalization,''
  \emph{Pattern Recognition}, 2020.

\bibitem{deng2020representation}
Z.~Deng, F.~Ding, C.~Dwork, R.~Hong, G.~Parmigiani, P.~Patil, and P.~Sur,
  ``Representation via representations: Domain generalization via adversarially
  learned invariant representations,'' \emph{arXiv:2006.11478}, 2020.

\bibitem{ganin2016domain}
Y.~Ganin, E.~Ustinova, H.~Ajakan, P.~Germain, H.~Larochelle, F.~Laviolette,
  M.~Marchand, and V.~Lempitsky, ``Domain-adversarial training of neural
  networks,'' \emph{JMLR}, 2016.

\bibitem{li2018deep}
Y.~Li, X.~Tian, M.~Gong, Y.~Liu, T.~Liu, K.~Zhang, and D.~Tao, ``Deep domain
  generalization via conditional invariant adversarial networks,'' in
  \emph{ECCV}, 2018.

\bibitem{arjovsky2019invariant}
M.~Arjovsky, L.~Bottou, I.~Gulrajani, and D.~Lopez-Paz, ``Invariant risk
  minimization,'' \emph{arXiv:1907.02893}, 2019.

\bibitem{blanchard2017domain}
G.~Blanchard, A.~A. Deshmukh, U.~Dogan, G.~Lee, and C.~Scott, ``Domain
  generalization by marginal transfer learning,'' \emph{arXiv:1711.07910},
  2017.

\bibitem{krueger2021out}
D.~Krueger, E.~Caballero, J.-H. Jacobsen, A.~Zhang, J.~Binas, D.~Zhang,
  R.~Le~Priol, and A.~Courville, ``Out-of-distribution generalization via risk
  extrapolation (rex),'' in \emph{ICML}, 2021.

\bibitem{zhang2020adaptive}
M.~Zhang, H.~Marklund, N.~Dhawan, A.~Gupta, S.~Levine, and C.~Finn, ``Adaptive
  risk minimization: A meta-learning approach for tackling group distribution
  shift,'' \emph{arXiv:2007.02931}, 2020.

\bibitem{li2018learning}
D.~Li, Y.~Yang, Y.-Z. Song, and T.~M. Hospedales, ``Learning to generalize:
  Meta-learning for domain generalization,'' in \emph{AAAI Conference on
  Artificial Intelligence}, 2018.

\bibitem{balaji2018metareg}
Y.~Balaji, S.~Sankaranarayanan, and R.~Chellappa, ``Metareg: Towards domain
  generalization using meta-regularization,'' \emph{NeurIPS}, 2018.

\bibitem{zhang2017mixup}
H.~Zhang, M.~Cisse, Y.~N. Dauphin, and D.~Lopez-Paz, ``mixup: Beyond empirical
  risk minimization,'' \emph{arXiv:1710.09412}, 2017.

\bibitem{xu2020adversarial}
M.~Xu, J.~Zhang, B.~Ni, T.~Li, C.~Wang, Q.~Tian, and W.~Zhang, ``Adversarial
  domain adaptation with domain mixup,'' in \emph{AAAI Conference on Artificial
  Intelligence}, 2020.

\bibitem{yan2020improve}
S.~Yan, H.~Song, N.~Li, L.~Zou, and L.~Ren, ``Improve unsupervised domain
  adaptation with mixup training,'' \emph{arXiv:2001.00677}, 2020.

\bibitem{wang2020heterogeneous}
Y.~Wang, H.~Li, and A.~C. Kot, ``Heterogeneous domain generalization via domain
  mixup,'' in \emph{IEEE International Conference on Acoustics, Speech and
  Signal Processing (ICASSP)}, 2020.

\bibitem{nam2019reducing}
H.~Nam, H.~Lee, J.~Park, W.~Yoon, and D.~Yoo, ``Reducing domain gap via
  style-agnostic networks,'' \emph{arXiv e-prints}, 2019.

\bibitem{goodfellow2014explaining}
I.~J. Goodfellow, J.~Shlens, and C.~Szegedy, ``Explaining and harnessing
  adversarial examples,'' \emph{arXiv:1412.6572}, 2014.

\bibitem{sinha2017certifying}
A.~Sinha, H.~Namkoong, R.~Volpi, and J.~Duchi, ``Certifying some distributional
  robustness with principled adversarial training,'' \emph{arXiv:1710.10571},
  2017.

\bibitem{volpi2018generalizing}
R.~Volpi, H.~Namkoong, O.~Sener, J.~Duchi, V.~Murino, and S.~Savarese,
  ``Generalizing to unseen domains via adversarial data augmentation,''
  \emph{arXiv:1805.12018}, 2018.

\bibitem{qiao2020learning}
F.~Qiao, L.~Zhao, and X.~Peng, ``Learning to learn single domain
  generalization,'' in \emph{IEEE/CVF CVPR}, 2020.

\bibitem{shankar2018generalizing}
S.~Shankar, V.~Piratla, S.~Chakrabarti, S.~Chaudhuri, P.~Jyothi, and
  S.~Sarawagi, ``Generalizing across domains via cross-gradient training,''
  \emph{arXiv:1804.10745}, 2018.

\bibitem{rahman2019multi}
M.~M. Rahman, C.~Fookes, M.~Baktashmotlagh, and S.~Sridharan, ``Multi-component
  image translation for deep domain generalization,'' in \emph{2019 IEEE Winter
  Conference on Applications of Computer Vision (WACV)}, 2019.

\bibitem{somavarapu2020frustratingly}
N.~Somavarapu, C.-Y. Ma, and Z.~Kira, ``Frustratingly simple domain
  generalization via image stylization,'' \emph{arXiv:2006.11207}, 2020.

\bibitem{borlino2021rethinking}
F.~C. Borlino, A.~D'Innocente, and T.~Tommasi, ``Rethinking domain
  generalization baselines,'' in \emph{International Conference on Pattern
  Recognition (ICPR)}, 2021.

\bibitem{maria2019hallucinating}
F.~Maria~Carlucci, P.~Russo, T.~Tommasi, and B.~Caputo, ``Hallucinating
  agnostic images to generalize across domains,'' in \emph{IEEE/CVF ICCV
  Workshops}, 2019.

\bibitem{zhou2020deep}
K.~Zhou, Y.~Yang, T.~Hospedales, and T.~Xiang, ``Deep domain-adversarial image
  generation for domain generalisation,'' in \emph{Proceedings of the AAAI
  Conference on Artificial Intelligence}, vol.~34, no.~07, 2020, pp.
  13\,025--13\,032.

\bibitem{zhou2020learning}
------, ``Learning to generate novel domains for domain generalization,'' in
  \emph{European conference on computer vision}.\hskip 1em plus 0.5em minus
  0.4em\relax Springer, 2020, pp. 561--578.

\bibitem{huang2020self}
Z.~Huang, H.~Wang, E.~P. Xing, and D.~Huang, ``Self-challenging improves
  cross-domain generalization,'' in \emph{Computer Vision--ECCV 2020: 16th
  European Conference}, 2020.

\bibitem{zhou2021mixstyle}
K.~Zhou, Y.~Yang, Y.~Qiao, and T.~Xiang, ``Domain generalization with
  mixstyle,'' \emph{arXiv:2104.02008}, 2021.

\bibitem{simonyan2014very}
K.~Simonyan and A.~Zisserman, ``Very deep convolutional networks for
  large-scale image recognition,'' \emph{arXiv:1409.1556}, 2014.

\bibitem{schulz2020restricting}
K.~Schulz, L.~Sixt, F.~Tombari, and T.~Landgraf, ``Restricting the flow:
  Information bottlenecks for attribution,'' \emph{arXiv:2001.00396}, 2020.

\bibitem{smilkov2017smoothgrad}
D.~Smilkov, N.~Thorat, B.~Kim, F.~Vi{\'e}gas, and M.~Wattenberg, ``Smoothgrad:
  removing noise by adding noise,'' \emph{arXiv:1706.03825}, 2017.

\bibitem{sundararajan2017axiomatic}
M.~Sundararajan, A.~Taly, and Q.~Yan, ``Axiomatic attribution for deep
  networks,'' in \emph{ICML}, 2017.

\bibitem{springenberg2014striving}
J.~T. Springenberg, A.~Dosovitskiy, T.~Brox, and M.~Riedmiller, ``Striving for
  simplicity: The all convolutional net,'' \emph{arXiv:1412.6806}, 2014.

\bibitem{bach2015pixel}
S.~Bach, A.~Binder, G.~Montavon, F.~Klauschen, K.-R. M{\"u}ller, and W.~Samek,
  ``On pixel-wise explanations for non-linear classifier decisions by
  layer-wise relevance propagation,'' \emph{PloS one}, 2015.

\bibitem{montavon2017explaining}
G.~Montavon, S.~Lapuschkin, A.~Binder, W.~Samek, and K.-R. M{\"u}ller,
  ``Explaining nonlinear classification decisions with deep taylor
  decomposition,'' \emph{Pattern Recognition}, 2017.

\bibitem{zhou2016learning}
B.~Zhou, A.~Khosla, A.~Lapedriza, A.~Oliva, and A.~Torralba, ``Learning deep
  features for discriminative localization,'' in \emph{IEEE CVPR}, 2016.

\bibitem{srivastava2014dropout}
N.~Srivastava, G.~Hinton, A.~Krizhevsky, I.~Sutskever, and R.~Salakhutdinov,
  ``Dropout: a simple way to prevent neural networks from overfitting,''
  \emph{JMLR}, 2014.

\bibitem{van2019disentangled}
S.~van Steenkiste, F.~Locatello, J.~Schmidhuber, and O.~Bachem, ``Are
  disentangled representations helpful for abstract visual reasoning?''
  \emph{arXiv:1905.12506}, 2019.

\bibitem{kingma2014adam}
D.~P. Kingma and J.~Ba, ``Adam: A method for stochastic optimization,''
  \emph{arXiv:1412.6980}, 2014.

\bibitem{he2016deep}
K.~He, X.~Zhang, S.~Ren, and J.~Sun, ``Deep residual learning for image
  recognition,'' in \emph{IEEE CVPR}, 2016.

\bibitem{russakovsky2015imagenet}
O.~Russakovsky, J.~Deng, H.~Su, J.~Krause, S.~Satheesh, S.~Ma, Z.~Huang,
  A.~Karpathy, A.~Khosla, M.~Bernstein \emph{et~al.}, ``Imagenet large scale
  visual recognition challenge,'' \emph{International journal of computer
  vision}, 2015.

\bibitem{ioffe2015batch}
S.~Ioffe and C.~Szegedy, ``Batch normalization: Accelerating deep network
  training by reducing internal covariate shift,'' in \emph{ICML}, 2015.

\bibitem{seo2020learning}
S.~Seo, Y.~Suh, D.~Kim, G.~Kim, J.~Han, and B.~Han, ``Learning to optimize
  domain specific normalization for domain generalization,'' in \emph{ECCV},
  2020.

\bibitem{fang2013unbiased}
C.~Fang, Y.~Xu, and D.~N. Rockmore, ``Unbiased metric learning: On the
  utilization of multiple datasets and web images for softening bias,'' in
  \emph{IEEE ICCV}, 2013.

\bibitem{venkateswara2017deep}
H.~Venkateswara, J.~Eusebio, S.~Chakraborty, and S.~Panchanathan, ``Deep
  hashing network for unsupervised domain adaptation,'' in \emph{IEEE CVPR},
  2017.

\bibitem{li2017deeper}
D.~Li, Y.~Yang, Y.-Z. Song, and T.~M. Hospedales, ``Deeper, broader and artier
  domain generalization,'' in \emph{IEEE ICCV}, 2017.

\bibitem{paszke2019pytorch}
A.~Paszke, S.~Gross, F.~Massa, A.~Lerer, J.~Bradbury, G.~Chanan, T.~Killeen,
  Z.~Lin, N.~Gimelshein, L.~Antiga \emph{et~al.}, ``Pytorch: An imperative
  style, high-performance deep learning library,'' \emph{Advances in neural
  information processing systems}, vol.~32, pp. 8026--8037, 2019.

\bibitem{moritz2018ray}
P.~Moritz, R.~Nishihara, S.~Wang, A.~Tumanov, R.~Liaw, E.~Liang, M.~Elibol,
  Z.~Yang, W.~Paul, M.~I. Jordan \emph{et~al.}, ``Ray: A distributed framework
  for emerging $\{$AI$\}$ applications,'' in \emph{13th $\{$USENIX$\}$
  Symposium on Operating Systems Design and Implementation ($\{$OSDI$\}$ 18)},
  2018, pp. 561--577.

\bibitem{liaw2018tune}
R.~Liaw, E.~Liang, R.~Nishihara, P.~Moritz, J.~E. Gonzalez, and I.~Stoica,
  ``Tune: A research platform for distributed model selection and training,''
  \emph{arXiv preprint arXiv:1807.05118}, 2018.

\end{thebibliography}

\end{document}